\documentclass{egpubl}
\usepackage{pg2022}

\SpecialIssueSubmission    

\usepackage[T1]{fontenc}
\usepackage{dfadobe}  

\usepackage{cite}  
\BibtexOrBiblatex
\electronicVersion
\PrintedOrElectronic
\ifpdf \usepackage[pdftex]{graphicx} \pdfcompresslevel=9
\else \usepackage[dvips]{graphicx} \fi

\usepackage{egweblnk} 
\usepackage{booktabs}
\usepackage{amsmath}
\usepackage{hyperref}
\usepackage{amsfonts}

\title[SPCNet: Stepwise Point Cloud Completion Network]%
      {SPCNet: Stepwise Point Cloud Completion Network}

\author[Fei Hu et al.]
{\parbox{\textwidth}{\centering 
        Fei Hu$^{1,2}$\orcid{0000-0002-5468-3383},
        Honghua Chen$^{1}$,
        Xuequan Lu$^{3}$,
        Zhe Zhu$^{1,2}$,
        Jun Wang$^{1}$,
        Weiming Wang$^{4}$,
        Fu Lee Wang$^{4\dag}$,
        Mingqiang Wei$^{1,2}$\thanks{Co-corresponding authors: F. L. Wang and M. Wei.}
        }
        \\
{\parbox{\textwidth}{\centering $^1$School of Computer Science and Technology, Nanjing University of Aeronautics and Astronautics, Nanjing, China\\
         $^2$Shenzhen Research Institute, Nanjing University of Aeronautics and Astronautics, Shenzhen, China\\
         $^3$School of Information Technology, Deakin University, Geelong, Australia\\
         $^4$School of Science and Technology, Hong Kong Metropolitan University, Hong Kong, China
       }
}
}

\begin{document}


\maketitle
\begin{abstract}
How will you repair a physical object with large missings? You may first recover its global yet coarse shape and stepwise increase its local details. We are motivated to imitate the above physical repair procedure to address the point cloud completion task. 
We propose a novel stepwise point cloud completion network (SPCNet) for various 3D models with large missings. SPCNet has a hierarchical bottom-to-up network architecture. It fulfills shape completion in an iterative manner, which 1) first infers the global feature of the coarse result; 2) then infers the local feature with the aid of global feature; and 3) finally infers the detailed result with the help of local feature and coarse result. 
Beyond the wisdom of simulating the physical repair, we newly design a cycle loss 
to enhance the generalization and robustness of SPCNet. 
Extensive experiments clearly show the superiority of our SPCNet over the state-of-the-art methods on 3D point clouds with large missings.
Code is available at \url{https://github.com/1127368546/SPCNet}.


\end{abstract}  

\section{Introduction}
Point clouds captured by LiDAR scanners or depth cameras are often incomplete, due to the measurement and reconstruction errors, as well as the occlusions of objects. 
When utilizing these untreated point clouds for semantic tasks like object classification \cite{qi2017pointnet,han20193dviewgraph}, segmentation \cite{liu2019point2sequence,wen2020cf}, and shape retrieval \cite{han2019view,han2019parts4feature,han2018seqviews2seqlabels}, we may receive inaccurate or even wrong results. 
Therefore, point cloud completion is required which aims to infer the whole underlying surface from a partial observation. Moreover,
the completion results should be uniform, dense and possess topologically correct geometric structures.

Point cloud completion has attracted increasing attention in recent years. 
Conventional shape completion methods \cite{sung2015data,nguyen2016field,hu2019local} utilize the geometric information (e.g., non-local similarity, symmetry, contour) to infer the small missing parts, which generally requires manual prior knowledge in advance. However, for these conventional
wisdom of point cloud completion, users have to tune parameters 
multiple times to obtain relatively satisfied completion results for an input with the large part missing. This heavily discounts the efficiency and user experience.

Deep learning has been widely used for point cloud analysis and processing. Researchers have  contributed three strands of deep learning methods to point cloud completion: voxelization  \cite{han2017high,dai2017shape},  PointNet/PointNet++  \cite{yuan2018pcn,8990130} and graph-based methods \cite{9093117}.
They take a partial point cloud as input to various encoders directly, and the extracted features usually contain both the global shape and local detail information. After the encoder, some completion methods infer the completed shape by directly decoding the extracted features, which will output coarse completion results without well-recovered local geometry details; other works \cite{yuan2018pcn} further transform the coarse result to a detailed result by shared MLPs. 
Although higher quality results can be produced by this coarse-to-fine strategy, these methods essentially do not make full use of the global and local information. 
To address this problem, PF-Net \cite{huang2020pf} is designed as a multi-scale network, which can extract global and local features respectively, but these features are still under-explored. 

\begin{figure}[!ht]
\centering
\includegraphics[width=1\columnwidth]{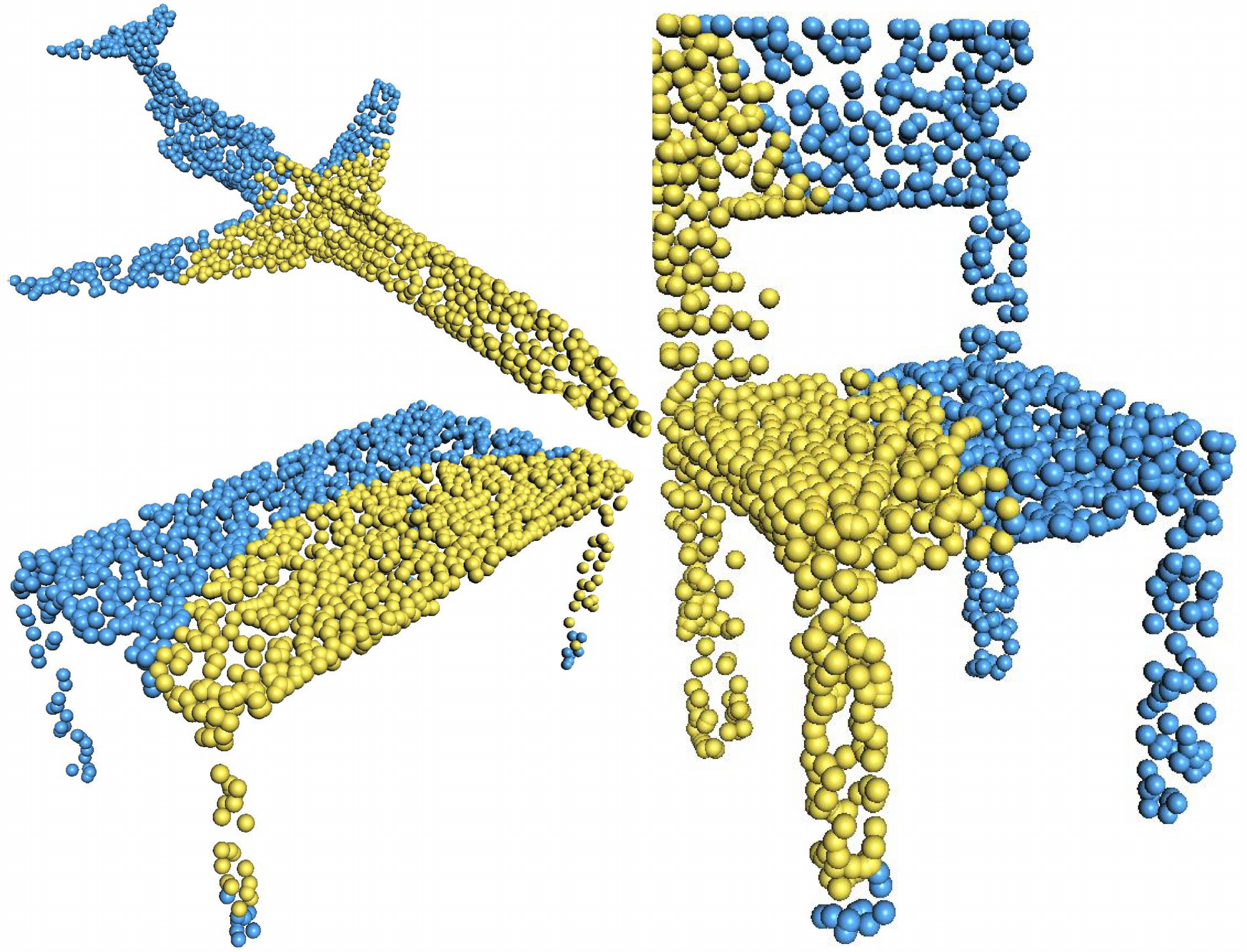} 
\caption{Shape completion for models with large-scale structure missings by SPCNet. Blue points belong to the original incomplete point clouds. Red points belong to the recovered shape. }
\label{completion}
\end{figure}

To desirably complete visual 3D models, one can absorb the wisdom of physical object repair. Imagine how the professional restorer repairs a physical object with large missings. She/he may first recover its coarse yet global shape and stepwise increase its local details. 
Thus, we attempt to imitate the aforementioned physical repair procedure to boost the performance of cutting-edge point cloud completion models.
We propose an effective stepwise point cloud completion network (SPCNet) for various 3D models with large missings. 
In order to imitate the repair procedure in SPCNet, we design SCM (stepwise completion module) and use it to generate better result iteratively. 
In SCM, we design VMLP (a variant of MLP) and ACM (Adaptive Convolution Module) to extract global and local features, respectively.
SCM will i) first infer the global feature by VMLP; ii) then infer the local feature by ACM; and iii) finally generate better result by the local feature.
Several shape completion example by SPCNet are shown in Fig. \ref{completion}, where large missings are recovered effectively. 

To make SPCNet `become' a professional restorer, i) we down-sample the input twice to obtain its multi-resolution representations, on which different-scale features can be extracted respectively; ii) we iteratively use SCM to stepwise repair the missing parts by utilizing the above multi-resolution representations; and iii) by optimizing the extra loss from the median results of SCM in the training stage, SCM will be more accurate in completion. 

Moreover, unlike existing completion methods that directly utilize the Chamfer distance loss, we introduce a cycle loss to optimize SPCNet based on CycleGan \cite{zhu2017unpaired}. 
SPCNet will output the missing parts of an original incomplete point cloud, then take the output as input again to receive the cycle-result, and finally calculate the cycle loss between the original input and the cycle-result. 
By using the cycle loss rather than the Chamfer distance loss, we can represent the structure difference between the inferred result and the ground truth more effectively.

The main contributions are as follows. 
\begin{itemize}
\item By imitating the physical repair procedure, we design SCM (stepwise completion module). SCM  transforms a coarse result to its fine version iteratively. 
\item We design the VMLP module (a variant of MLP) in SCM, which can aggregate the last global feature of SCM to the current feature to fully utilize the geometrical information.
\item We embed an adaptive convolution module in SCM, which can transform the global feature to the point-wise local feature, for generating a detailed completion result. 
\item We train the network with the proposed cycle loss to enhance the generalization and robustness of SPCNet, especially when handling point clouds with large missings. 
\end{itemize}

\section{Related Work}
 3D shape completion can be classified into two types:  traditional and deep learning methods.  Traditional methods can be further divided into  geometry-based and alignment-based methods. Also, there exist supervised and unsupervised methods for deep learning based techniques.

\subsection{Traditional methods}

Geometry-based methods \cite{sung2015data} utilize geometric information from the partial input to generate the complete shape. 
Some methods \cite{berger2014state,nguyen2016field,hu2019local} fill the missing parts locally by interpolating smooth surfaces from their adjacent structures. However, they cannot receive desired results when input models have large missing regions due to the lack of semantic information (e.g., structure, topology). Based on the observation that many objects are symmetric, some methods \cite{mitra2006partial,pauly2008discovering,thrun2005shape} can detect the symmetry in input 3D models, and fill the missings by the help of their symmetric geometry. 

Alignment-based methods \cite{martinovic2013bayesian,shao2012interactive,shen2012structure} take advantage of a large-scale point cloud database to search for the most suitable patches, and then fill the missing regions with these patches. The most important step in these methods is to create a database that  represents the input 3D shapes in the same category. Some methods \cite{kalogerakis2012probabilistic,kim2013learning,pauly2005example} choose parts of models as elements for the database, other methods \cite{chauve2010robust,rock2015completing,schnabel2009completion,yin2014morfit} choose the deformation of 3D shapes such as planes and quadrics. This way can complete 3D models that have similar structures, but usually fail when the 3D models are beyond the capability of the predefined categories.

In summary, these traditional methods depend on hand-crafted geometric features to infer the structure of missing regions from a partial shape. They barely receive satisfactory results when encountering 3D models that violate the prior knowledge. 

\begin{figure*}[!ht]
\centering
\includegraphics[width=0.8\linewidth]{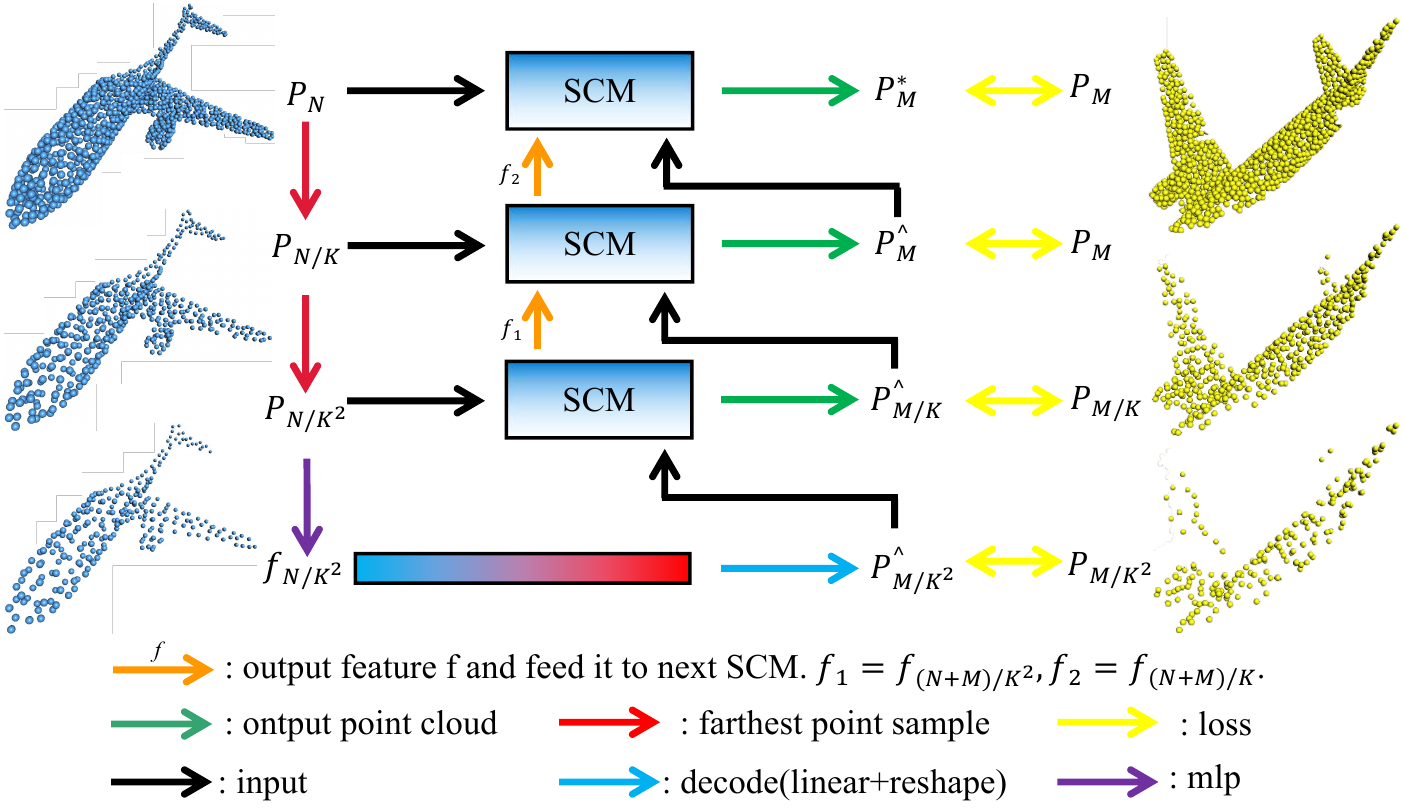} 
\caption{Pipeline of Stepwise Point Cloud Completion Network (SPCNet).
}
\label{net}
\end{figure*}

\subsection{Deep learning-base methods}

Supervised methods learn the shape representation from a partial input with an encoder, and predict the complete shape from this representation with a corresponding decoder. The so-called encoder-decoder network is optimized with paired training data. 
As for shape representation, some methods  \cite{han2017high,dai2017shape,xie2020grnet} leverage 3D volumetric convolution, and others \cite{yang2018foldingnet,sarmad2019rl,hu2019render4completion,wang2020cascaded} may borrow PointNet/PointNet++ to directly use point clouds as input.  
GR-Net \cite{xie2020grnet} designs a novel way to achieve the voxelization of point clouds, and obtains the best result in current 3D convolution completion methods. 
Employing 3D convolution for 3D shape completion is intuitive and convenient, it will probably receive better results if there exists a high-resolution 3D voxelization database and adequate computing capability.  
PCN \cite{yuan2018pcn} exploits a coarse-to-fine strategy, which directly transforms the input point cloud to a coarse result and then transforms the coarse result to a detailed result. Render4Completion \cite{hu2019render4completion} introduces a new loss function, which renders both the result and ground truth to 2D RGB images and calculates a 2D loss by these images. 
MSN \cite{liu2020morphing} is a novel folding algorithm, it can recover more detailed structures by transforming 2D grids to different 3D surfaces. 

In order to increase the learning effect when directly using point clouds as input, some works adopt graph convolution to point clouds. 
Kipf et al. \cite{kipf2016semi} propose a graph convolution network which carries out a weighted convolution operation over graphs that is built on adjacent nodes' features.  The variants of graph convolution \cite{wang2019dynamic,li2018pointcnn,shen2018mining,hua2018pointwise,lei2020spherical} are proposed to change the convolution weight or graph. DGCNN \cite{wang2019dynamic} creates the EdgeConv operator for feature extraction, which gathers neighboring points to the specific feature space. Some methods \cite{wang2019graph,verma2018feastnet,velivckovic2017graph} fuse the attention mechanism to graph convolution, and can generate attentional weights for points on a graph. In order to take better advantage of point cloud's shift and scale-invariant properties, Lin et al. \cite{lin2020convolution} propose the deformable graph convolution kernels to extract features. ECG-Net \cite{9093117} designs a new network like DGCNN \cite{wang2019dynamic} for point cloud completion, and receives promising completion results.  

Unsupervised methods utilize unpaired training data to learn the relationship between complete and incomplete shapes in the latent space. AML \cite{stutz2018learning} estimates the amortized maximum likelihood between the latent representation of complete and incomplete shapes. Pcl2Pcl \cite{chen2019unpaired} takes advantage of GAN to learn the semantic relationship from the distribution between complete and incomplete shapes.  Cycle4Completion  \cite{wen2021cycle4completion} utilizes a new way learned from CycleGan, which designs two generators and two discriminators to learn the bidirectional geometric correspondence between the latent space of complete and incomplete shapes. 

In short, supervised methods usually have better performance than unsupervised methods if a high-quality training database is available. The unsupervised methods are more valuable in practical situations where we cannot acquire adequate training data.  

Shape completion is an ill-conditioned problem in that the valid complete results for a same incomplete input are multiple and even countless. The reasons may be that 1) there are many different ways to sample the missing parts on the original 3D model, and 2) the missing parts of 3D model have many semantic possibilities. The proposed SPCNet will fit the spatial structure distribution of training point clouds after supervised learning, and could infer a complete result that follows this distribution.

\section{Method}
SPCNet is an end-to-end network with full supervision. It takes an incomplete point cloud as input, and outputs the missing parts instead of the whole shape, which can reduce the model size of our SPCNet and increase the training efficiency. 

We show the pipeline of SPCNet in Fig. \ref{net}. SPCNet takes an incomplete point cloud $P_N$ as input, and outputs $P_{M}^{*}$ as the final completion result.
$P_{M}^{*}$ is a uniform and dense point cloud that represents the missing part(s), and we assume ${P_N}\cup{P_M}$ is the whole ground-truth model of input $P_N$.
We first use the farthest point sampling (FPS) algorithm to sample $P_N$ with the rates of $K$ and $K^2$ respectively to obtain the down-sampled point clouds $P_{N/K}$ and $P_{N/{K^2}}$. Then, we use MLP to encode $P_{N/{K^2}}$ and get the feature $f_{N/{K^2}}$. $f_{N/{K^2}}$ is further fed into a decoder that consists of linear and reshape operations to generate the coarse result $P_{M/{K^2}}^{\hat{}}$. Here, we utilize SCM to iteratively generate a better shape. 
We first feed $P_{N/{K^2}}$ and $P_{M/{K^2}}^{\hat{}}$ into SCM, which will produce a better result $P_{M/K}^{\hat{}}$ and the point-wise global feature $f_{(N+M)/{K^2}}$. Then, we feed $P_{N/K}$, $P_{M/K}^{\hat{}}$ and $f_{(N+M)/{K^2}}$ into SCM, which will produce a better result $P_{M}^{\hat{}}$ and the feature $f_{(N+M)/K}$. Finally, we feed $P_N$, $P_{M}^{\hat{}}$ and $f_{(N+M)/K}$ to SCM and obtain the final result $P_{KM}^{\hat{}}$ (also named $P_{M}^{*}$). 
Therefore, the whole completion result is ${P_N}\cup{P_{M}^{*}}$. 
In addition, we down-sample the ground truth in the same way, and calculate loss functions between each result and its ground truth.

\begin{figure}[!ht]
\centering
\includegraphics[width=1\columnwidth]{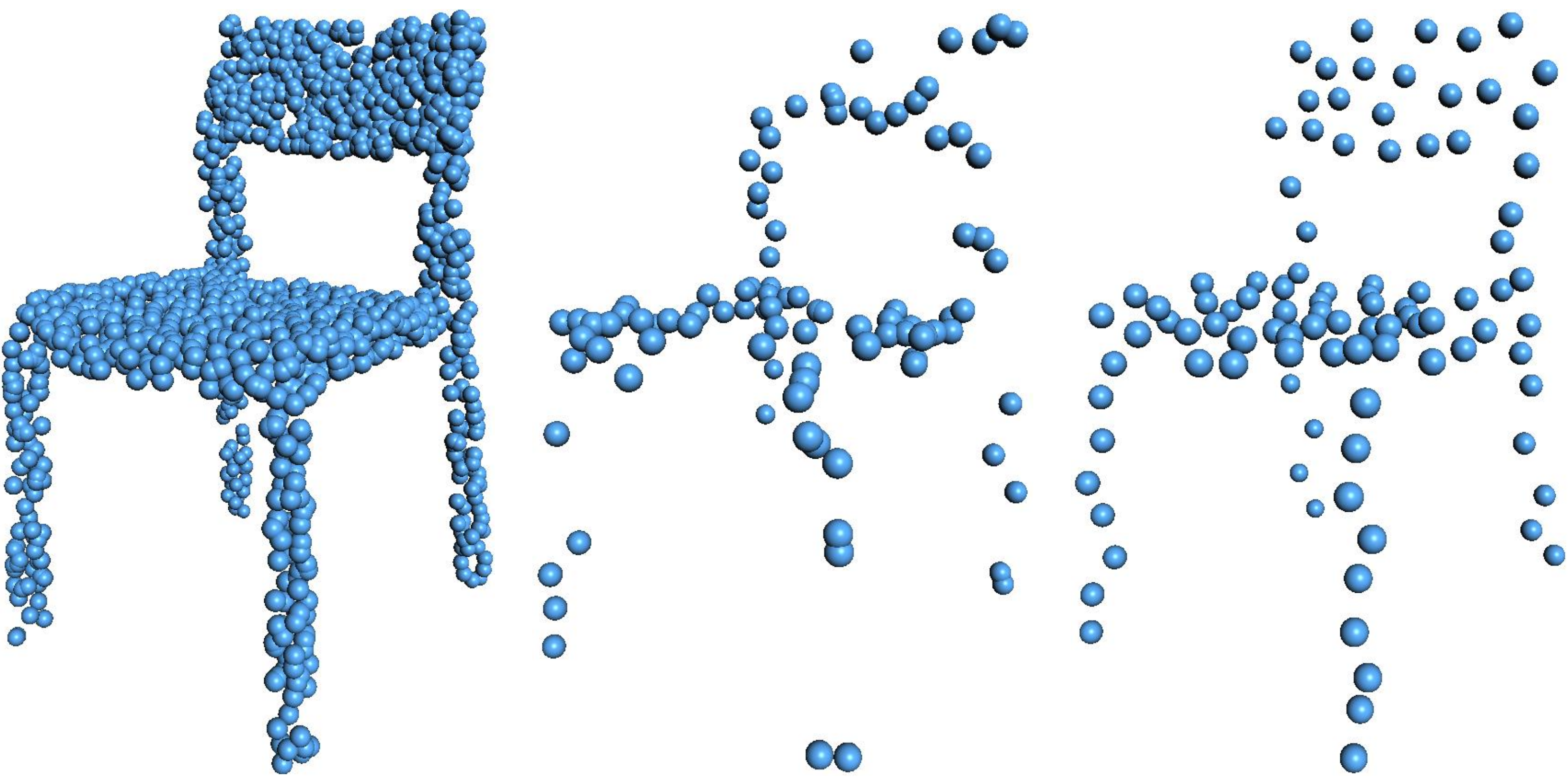} 
\caption{Left to right: the original object, the random sampling result, and the farthest point sampling result. }
\label{fps}
\end{figure}

\subsection{Multi-scale representations}
To extract adequate geometric information from latent space, we first generate multi-resolution representations of the input point cloud by multiple down-sampling operations. After that, we obtain more global information from the sparser representation and more local information from the denser representation. 
Random sampling is the fastest solution, but may lose some important `skeleton' points. Also, there are some down-sampling based networks achieving excellent results, but we do not intend to employ them because of the time-consuming calculation. 
We trade off effectiveness and efficiency by choosing the farthest point sampling (FPS) scheme \cite{NIPS2017_d8bf84be}.
FPS iteratively selects the most distant point as sampled point with regard to the rest points.
From Fig. \ref{fps} it can be observed that the FPS result maintains the integral structure of the input point cloud even if it only involves 5\% points of the input.

\subsection{Stepwise Completion Module}
We design SCM to transform the coarse result to its fine version, then we can stepwise complete the missing parts. 
For example, our third (highest-resolution) SCM is shown in Fig. \ref{scm} (the others can be found in supplementary material), we take the partial input $P_N$, the coarse result $P_{M}^{\hat{}}$ and the last SCM's feature $f_{(N+M)/K}$ as input.
First, we concatenate $P_N$ and $P_{M}^{\hat{}}$, and receive a whole point cloud $P_{N+M}^{\hat{}}$.
Then, we input $P_{N+M}^{\hat{}}$ to VMLP, and receive the point-wise global feature $f_{N+M}^{\hat{}}$.
Next, we aggregate the feature $f_{(N+M)/K}$ to the feature $f_{N+M}^{\hat{}}$, receive the new point-wise global feature $f_{N+M}$, and feed it to the next SCM (if it exists).
Finally, we input $P_{M}^{\hat{}}$, $P_{N+M}^{\hat{}}$ and $f_{N+M}$ to ACM, and receive the detailed yet final completion result $P_{KM}^{\hat{}}$ (also named $P_{M}^{*}$).

\begin{figure}[!ht]
\centering
\includegraphics[width=1\columnwidth]{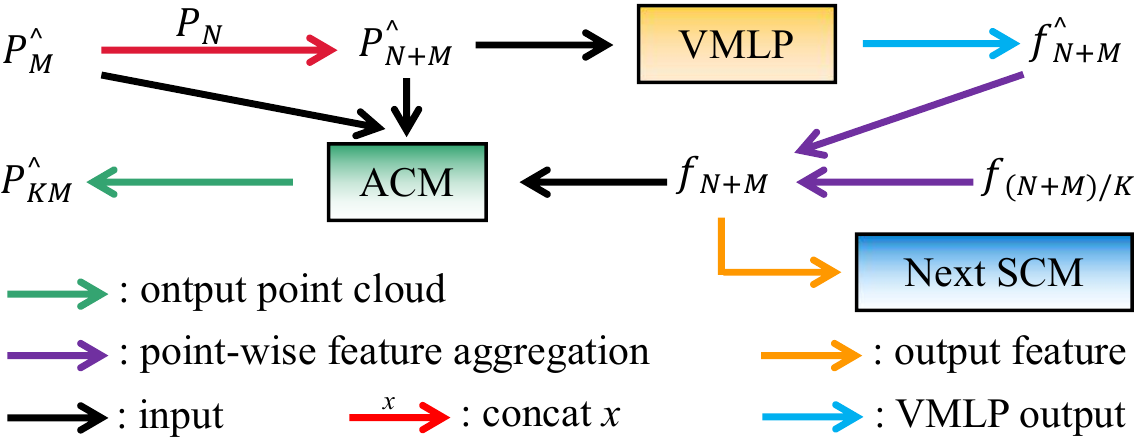} 
\caption{Third SCM (Stepwise Completion Module).}
\label{scm}
\end{figure}

Here, we explain the aggregation process. 
For each point $p_i$ in $P_{N+M}^{\hat{}}$ with the feature $f_{N+M}^{\hat{}}$, we search its closest point $p_j$ in  $P_{(N+M)/K}^{\hat{}}$, 
then concatenate this closest point's feature $f_{(N+M)/K}[j]$ to the $p_i$'s feature $f_{N+M}^{\hat{}}[i]$.
We apply a linear operation to concatenate the result, and receive an aggregated point-wise feature $f_{N+M}$.
Aggregating the last SCM’s global feature to the current feature, we can fully utilize the geometrical information from the aggregated spatial structure.

\subsection{Variant of MLP}
PointNet-MLP maps $N$ points into $N$ features with different dimensions (like [64, 128, 256, 512, 1024]) in each layer, then extracts the maximum value from columns of the last layer's $N$ features to form a global latent vector. However, it will waste much local information extracted in the median layers. Therefore we improve PointNet-MLP, called VMLP, to better extract the global information, and VMLP in our third SCM is shown in Fig. \ref{vmlp} (others can be found in the supplementary material). 

\textbf{VMLP}. 
In order to take better advantage of spatial geometric information from spatial coordinates, we design three MLP-Sub-Nets instead of single MLP-Net. 
At each MLP-Sub-Net, after maxpooling and concatenating the results of the last four layers, we can receive a feature that contains both low-level and high-level global information. Then, we apply a linear layer to adjust the concatenate results. 
Finally, we respectively repeat and concatenate three MLP-Sub-Net's final features to x, y and z coordinates, and apply adaptive graph convolution (we will explain later) to it, then receive the point-wise global feature.

\begin{figure}[!ht]
\centering
\includegraphics[width=1\columnwidth]{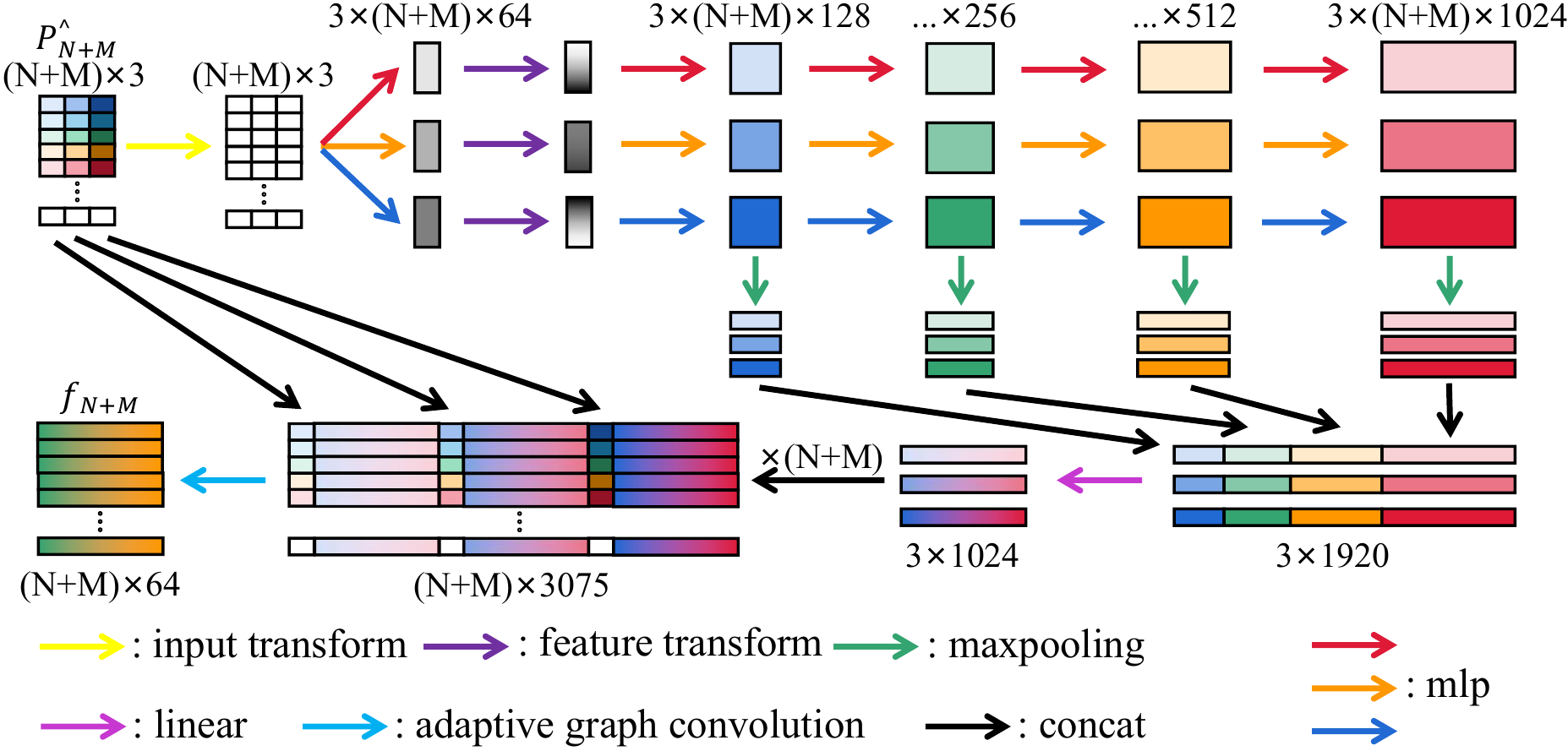} %
\caption{Network of VMLP (a variant of MLP) in our third SCM. The input transform and feature transform operations are same as PointNet \cite{qi2017pointnet}.
}
\label{vmlp}
\end{figure}

\subsection{Adaptive Convolution Module}

\begin{figure*}[t]
\centering
\includegraphics[width=1\textwidth]{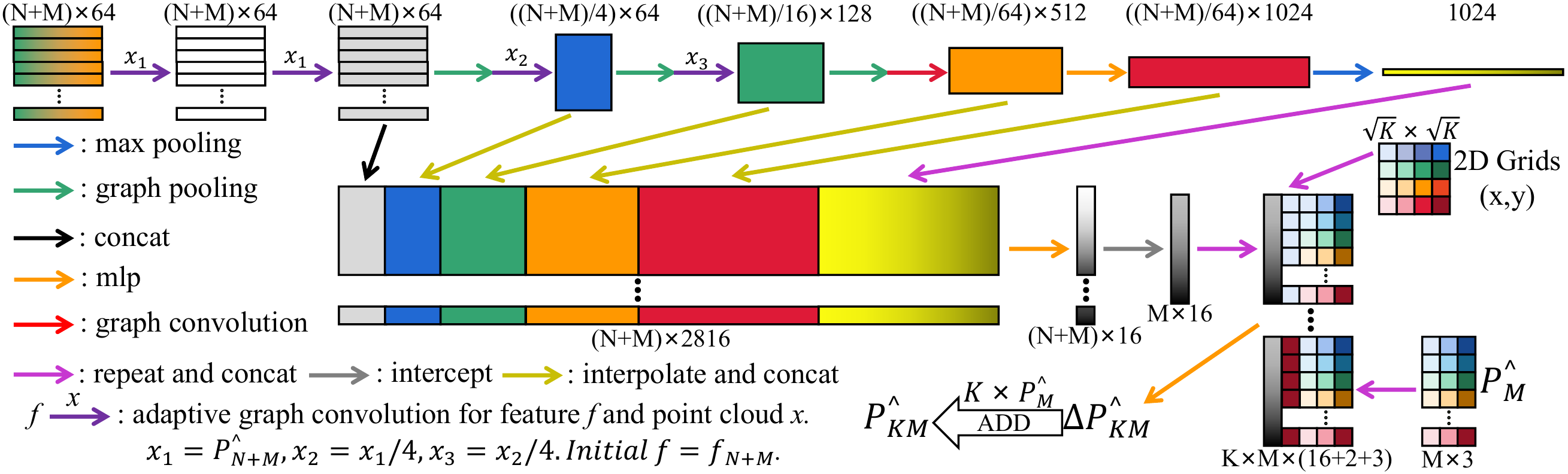} %
\caption{Network of ACM (Adaptive Convolution Module) in our third SCM. The graph convolution is same as \cite{wang2019dynamic}.
}
\label{adaptconv}
\end{figure*}

Despite that our VMLP can extract more information especially the global information, when it comes to the local information, we usually cannot receive tiptop results. It motivates us to design a new way to extract the local information. Here, we choose the AdaptConv (adaptive convolution \cite{zhou2021adaptive}) layer that contains adaptive graph convolution operations to extract the local information. ACM consists of an AdaptConv layer encoder and a decoder. ACM in our third SCM is shown in Fig. \ref{adaptconv} (others can be found in the supplementary material).

\textbf{AdaptConv layer encoder.} The graph convolution (including AdaptConv) extracts each point's feature from its neighboring points. If an incomplete point cloud $P_N$ is sent to AdaptConv directly, we may receive the feature with misleading information. 
To mitigate this, we choose to feed whole point cloud $P_{N+M}^{\hat{}}$ to AdaptConv layer. 
We adopt AdaptConv, graph convolution, and MLP to extract the local information for each point, and utilize graph pooling to expose more in-depth information by shrinking the points' number $N+M$, and finally reap a final local feature with max-pooling. In the last stage, we up-sample some intermediate features by interpolating neighboring points' features, and then concatenate an intermediate feature, some up-sampling features and the repeating final local feature, finally receiving a detailed point-wise local feature by feeding the concatenated result to MLP. We will explain AdaptConv and graph pooling in next subsection.  

\textbf{AdaptConv.} We assume that the input point cloud is $\mathcal{X} = \{x_i | i=1,2,...,N\} \in \mathbb{R}^{N \times 3}$, where $x_i$ represents the $(x,y,z)$ coordinates of the $i$-th point, and the corresponding feature is $\mathcal{F} = \{f_i | i=1,2,...,N\} \in \mathbb{R}^{N \times D}$. For each point, we need to generate an $M$-dimensional feature with its neighboring $D$-dimensional features. Previous graph convolution adopts the fixed kernel for each point, while AdaptConv generates adaptive kernels depending on the local information. Thus, we can more accurately extract the local spatial geometric information than previous graph convolution methods. 

We need to construct a directed graph $\mathcal{G}(\mathcal{V},\mathcal{E})$ from the given point cloud where $\mathcal{V} = \{1,...,N\}$ denotes the set of points, and $\mathcal{E} \subseteq \mathcal{V} \times \mathcal{V}$ denotes the set of edges, and we build its edge relation by utilizing its $K$ nearest neighbors.  $x_i$ is the current central point in AdaptConv, and we use $\mathcal{N}(i) = \{j : (i,j) \in \mathcal{E}\}$ as its neighborhood. For each channel in the output $M$-dimensional feature, AdaptConv dynamically generates a kernel using a function over the point coordinates $(x_i, x_j)$:
\begin{equation}
\hat{e}_{ijm} = g_m(\Delta x_{ij}), j \in \mathcal{N}(i),
\end{equation}
where $m = 1,2,...,M$ indicates one of the $M$ output dimensions corresponding to a single filter defined in AdaptConv.  $\Delta x_{ij} = [x_i, x_j-x_i]$, $[\cdot,\cdot]$ is the concatenation operation and $g(\cdot)$ is a feature mapping function consisting of MLP. 

Then, the adaptive kernel is convolved with the corresponding points' feature $(f_i, f_j)$:
\begin{equation}
h_{ijm} = \sigma \left \langle \hat{e}_{ijm}, \Delta f_{ij} \right \rangle, \label{equ:convolution}
\end{equation}
where $\langle \cdot, \cdot \rangle$ represents the inner product of two vectors outputting $h_{ijm} \in \mathbb{R}$ and $\sigma$ is a nonlinear activation function. Stacking  $h_{ijm}$ of each channel yields the edge feature $h_{ij} = [h_{ij1}, h_{ij2}, ..., h_{ijM}] \in \mathbb{R}^M$ between the connected points $(x_i, x_j)$.

Finally, applying an aggregating function over all the features in the neighborhood, we can receive the output feature of the current central point $x_i$:
\begin{equation}
f_i' = \max_{j \in \mathcal{N}(i)} h_{ij},
\end{equation}
where $\max$ is a channel-wise max-pooling function. The convolution weights of AdaptConv are defined as $\Theta = (g_1, g_2, ..., g_M)$. 

\textbf{Graph pooling.} For extracting the local feature at a deeper level, we adopt graph pooling to reduce the number of points progressively by constructing a hierarchical architecture. We receive a sub-point-cloud by utilizing farthest point sampling on this layer's input point cloud. Then, we can obtain each sub-point-cloud points' feature by applying an AdaptConv layer to aggregate the neighboring points' features. By this means, we can receive a sub-point-cloud with deeper features, and construct a new smaller graph in the following graph convolution or AdaptConv. 

\textbf{ACM decoder.}
After applying the AdaptConv layer encoder to $P_{N+M}^{\hat{}}$ and $f_{(N+M)/K}$, we receive a point-wise local feature for $P_{N+M}^{\hat{}}$.
First, we intercept the point-wise local feature from the dimension $N+M$ to $M$, and let the new point-wise local feature correspond to $P_{M}^{\hat{}}$.
Then, we sample 2D grids from a square plane ($r=0.05$).
Next, we concatenate 2D grids, the intercepted feature and $P_{M}^{\hat{}}$.
After applying MLP to concatenate the result, we receive $K$ displacements for $P_{M}^{\hat{}}$.
Finally, we receive the detailed result $P_{KM}^{\hat{}}$ (also named $P_{M}^{*}$) by adding these displacements to $K$ 
$P_{M}^{\hat{}}$.

\subsection{Cycle Training}
The incomplete input $P_N$ is sampled from a complete 3D model. But $P_N$ loses the part that is denoted by the point cloud $P_M$. We aim to yield $P_M$ to constitute the complete result ${P_N}\cup{P_M}$.  

For optimization, we need to ensure that ${P_N}\cup{P_{M}^{*}}$ is approximate to ${P_N}\cup{P_M}$, i.e., $P_{M}^{*}$ is close to $P_M$. As shown in Fig. \ref{net}, to achieve this, we narrow the gap progressively by employing the FPS scheme on $P_M$ to get $P_{M/K}$ and $P_{M/{K^2}}$, and make sure that $P_{M/K}^{\hat{}}$, $P_{M/{K^2}}^{\hat{}}$ are close to $P_{M/K}$, $P_{M/{K^2}}$, respectively. To be specific, we calculate four loss functions between them and add these results together. In this way, we can get more detailed results step by step, and finally make the distribution of $P_{M}^{\hat{}}$ and $P_{M}^{*}$ close to $P_M$.

To measure the spatial structure similarity between two point clouds $A$ and $B$, the Chamfer distance (CD) is formulated as: 
\begin{equation}
\begin{aligned}
    d_{CD}(A,B)=\frac{1}{|A|}\sum_{x\in A}\min_{y\in B}||x-y||_{2}^{2}
    \\+\frac{1}{\left | B \right |}\sum_{x\in B}\min_{y\in A}||x-y||_{2}^{2}.
\end{aligned}
\end{equation}

The overall loss function of SPCNet is expressed as:
\begin{equation}
\begin{split}
     Loss(P_{M}^{*},P_M)&=\alpha _1d_{CD}(P_{M/{K^2}^{\hat{}}},P_{M/{K^2}})+\alpha _2d_{CD}(P_{M/K}^{\hat{}},P_{M/K})\\
                        &+\alpha _3d_{CD}(P_{M}^{\hat{}},P_M)+\alpha _4d_{CD}(P_{M}^{*},P_M).
\end{split}
\end{equation}

The CD loss takes the least time in all shape completion losses, receiving a desirable result. Nevertheless, it cannot entirely measure the similarity between the ground truth and the completion result.  For example, if we repeatedly sample a point cloud from the same 3D model, we can obtain different point sets, but the CD between these sets in a same 3D model are not zero. 
In this work, we propose a new loss called cycle loss to enhance the structure similarity measurement ability of the original CD loss. 

\begin{figure}[!ht]
\centering
\includegraphics[width=1\columnwidth]{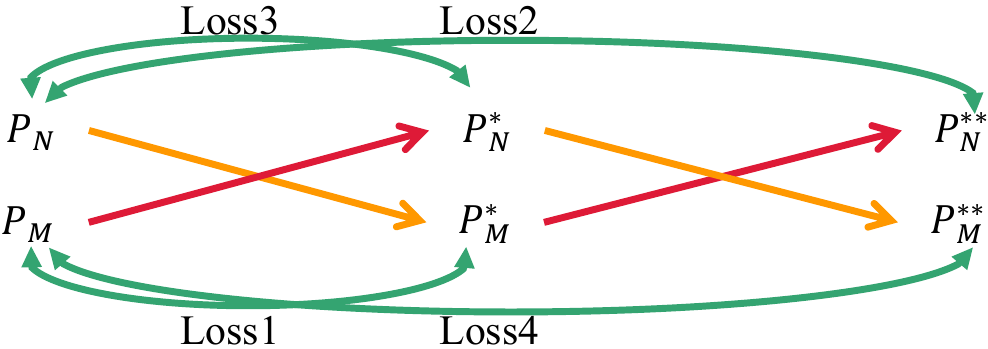} 
\caption{Our proposed cycle loss strategy. The point cloud in the same row has an equal spatial structure. Each complete point cloud is obtained by merging the two point clouds in the same column. }
\label{cycleloss}
\end{figure}

Fig. \ref{cycleloss} shows our whole training strategy. First, we separate the complete point cloud into two parts, one is ${P_N}$ and the other is ${P_M}$. 
We individually take ${P_N}$ and ${P_M}$ as input to our SPCNet, then receive ${P_{M}^{*}}$ and ${P_{N}^{*}}$ as output.
By repeating the same operation by taking ${P_{N}^{*}}$ and ${P_{M}^{*}}$ as input, we finally receive ${P_{M}^{**}}$ and ${P_{N}^{**}}$ as output. If the network is ideal,  ${P_N}$, ${P_{N}^{*}}$ and ${P_{N}^{**}}$ should be all equal, as well as ${P_M}$, ${P_{M}^{*}}$ and ${P_{M}^{**}}$. 
We optimize the network by calculating multiple losses between them. We calculate four losses, including two direct losses and two cycle losses. 
By taking the original input to the network, we receive the first output. Then, by taking the first output to the network, we can reap the final output, and we can calculate the CD loss between the original input and the final output. This cycle manner has the following benefits: (1) it can extract and measure more structure information than a single CD loss; (2) it can better reduce the structure difference between the network's output and the ground truth; and (3) it can increase the network's generalization ability. 

As a result, the total loss function is formulated as (the value of $\alpha$ and $\beta$ in training can be found in the supplementary material):
\begin{equation}
\begin{aligned}
    L_{total}=\beta_1 Loss(P_{M}^{*},P_M)+\beta_2 Loss(P_{N}^{**},P_N)
    \\+\beta_1 Loss(P_{N}^{*},P_N)+\beta_2 Loss(P_{M}^{**},P_M).
\end{aligned}
\end{equation}

\begin{table*}[!ht]
\centering
\footnotesize
\begin{tabular}{cccccccccc}
\toprule[1pt]
Category   & LGAN-AE        & PCN   & 3D-Capsule & TopNet   & MSN         & PF-Net          &ECG-Net &VRCNet           & SPCNet (Ours)          \\
\hline\hline
Airplane   & 2.814          & 2.626 & 2.991      & 2.251 & 1.698          & \textbf{0.984}  & 1.095  & 1.107           & 1.014                  \\
Bag        & 8.837          & 8.673 & 8.492      & 7.887 & 9.745          & 3.543           & 3.995  & 3.640           & \textbf{3.428}         \\
Cap        & 7.609          & 7.126 & 7.706      & 6.524 & 5.491          & 5.473           & 4.668  & 3.939           & \textbf{3.267}         \\
Car        & 5.416          & 5.789 & 6.236      & 5.514 & 5.716          & 2.390           & 2.496  & 2.312           & \textbf{2.281}         \\
Chair      & 4.787          & 4.153 & 4.045      & 3.597 & 3.072          & 2.053           & 2.124  & 2.111           & \textbf{1.835}         \\
Guitar     & 1.251          & 1.113 & 1.294      & 0.976 & 0.836          & 0.407           & 0.478  & 0.525           & \textbf{0.392}         \\
Lamp       & 7.476          & 6.918 & 7.669      & 6.534 & 3.517          & 4.185           & 3.467  & \textbf{2.709}  & 2.714                  \\
Laptop     & 3.376          & 3.262 & 3.627      & 2.671 & 1.619          & 1.448           & 1.408  & \textbf{1.215}  & 1.306                  \\
Motorbike  & 4.156          & 4.012 & 4.048      & 3.546 & 2.963          & 1.923           & 2.034  & 2.240           & \textbf{1.893}         \\
Mug        & 6.516          & 6.845 & 7.051      & 6.781 & 8.795          & 3.377           & 3.775  & \textbf{2.874}  & 2.976                  \\
Pistol     & 3.261          & 3.163 & 3.212      & 2.620 & 1.647          & 1.381           & 1.237  & 1.513           & \textbf{1.124}         \\
Skateboard & 3.022          & 2.906 & 3.346      & 2.717 & 1.760          & 1.327           & 1.354  & \textbf{1.115}  & 1.206                  \\
Table      & 4.781          & 4.746 & 5.157      & 4.036 & 4.342          & 2.053           & 1.982  & 1.983           & \textbf{1.867}         \\
\hline
Mean       & 4.869          & 4.717 & 4.990      & 4.281 & 3.938          & 2.349           & 2.316  & 2.098           & \textbf{1.946}         \\
\bottomrule[1pt]
\end{tabular}
\vspace{2pt}
\caption{Quantitative results. We calculate the CD loss of 13 categories from the test set, and the last row  denotes the mean CD loss of these categories. 
We calculate these results with the whole model between ${P_N}\cup{P_{M}^{*}}$ and ${P_N}\cup{P_M}$, scaled by 1000. }
\label{t1}
\end{table*}

\section{Experiments}

\subsection{Data Generation and Model Training}
We train and evaluate SPCNet on the benchmark dataset ShapeNet-Part, which has 13 categories of different objects. 
The dataset has 14,473 shapes formatted for point clouds, including 11,705 point clouds for training and 2,768 for testing. In this dataset, all point clouds are centered at the origin, i.e., their point coordinates are located within $[-1,1]$.  
We sample 2,048 points uniformly from each point cloud as a complete shape ${P_N}\cup{P_M}$. We select some border points like $(1,1,1)$ as viewpoints and randomly choose a viewpoint in each training epoch, then separate a certain amount of points close to the viewpoint, to get two point clouds ${P_N}$ and ${P_M}$. By changing the ratio of ${P_M}$ to ${P_N}\cup{P_M}$, we can control the missing rate. In our experiment, we are looking at the large-ratio incomplete point cloud completion problem. Thus, we set the ratio to 50\% for training and testing. Finally we will test other ratios for the robustness experiment. 

We design and train our network on PyTorch. All networks are optimized by an ADAM optimizer, with a learning rate of 0.0001 and a batch size of 24. We apply RELU activation units and batch normalization after each convolution, except for the last layer. In experiments, we set the down-sampling rate $K$ to 4, and 2D grid number to 16. $K$ in the first and second SCM is set to 4, and 1 in the last SCM.  

\begin{figure*}[!ht]
\centering
\includegraphics[width=0.84\textwidth]{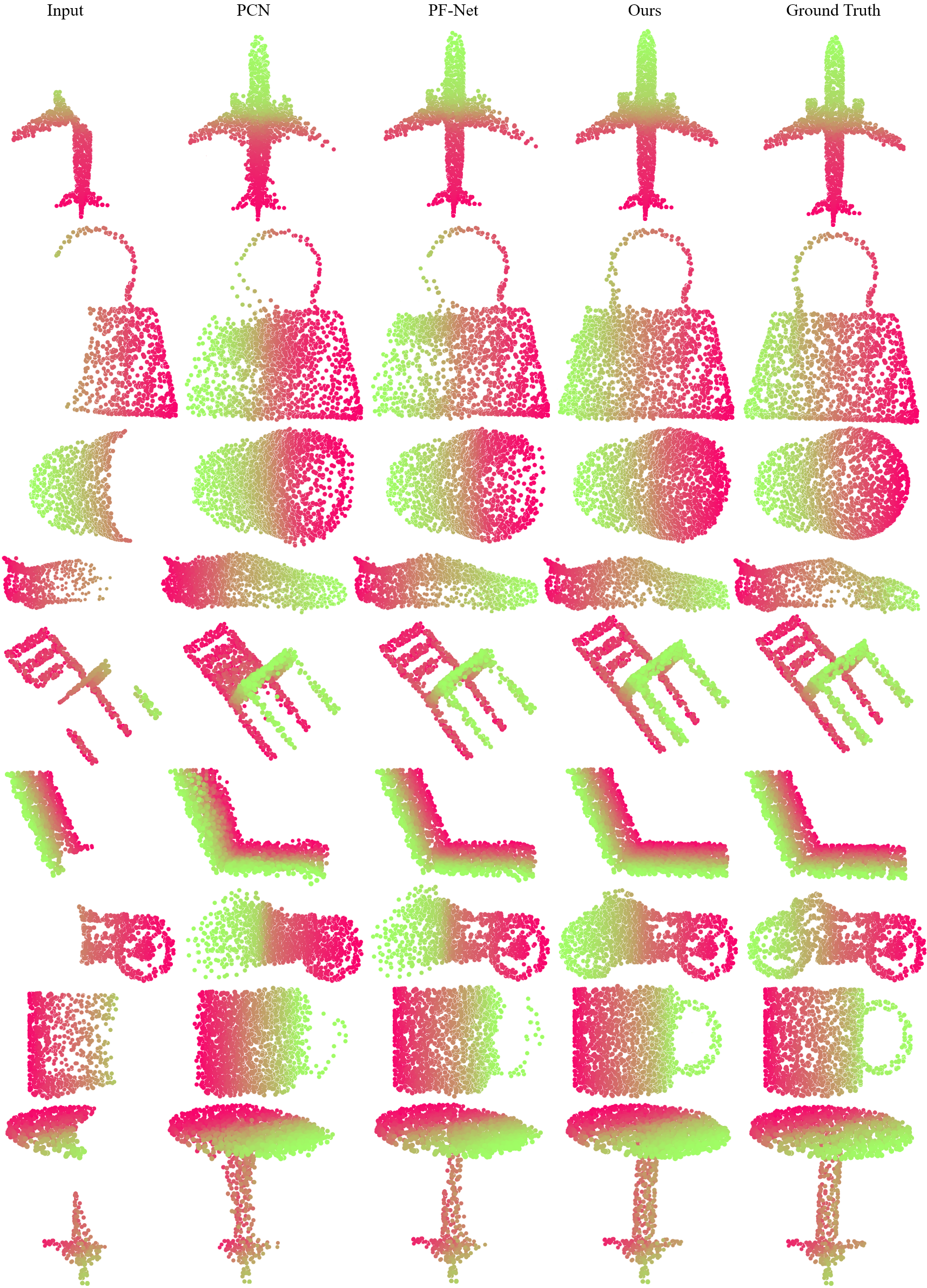} 
\caption{Visual comparisons. We compare our method with PCN, PF-Net, in 9 categories, including Airplane, Bag, Cap, Car, Chair, Laptop, Motorbike, Mug and Table. 
Our method has the best visual performance on average, which corresponds to the quantitative results. }
\label{compare}
\end{figure*}

\subsection{Completion Results}
We compare our method against several representative competitors including L-GAN \cite{lin2018learning}, PCN \cite{yuan2018pcn}, 3D-Capsual \cite{zhao20193d}, TopNet \cite{tchapmi2019topnet}, MSN \cite{liu2020morphing}, PF-Net \cite{huang2020pf}, ECG-Net \cite{9093117} and VRCNet \cite{9577912}. Our network is bigger than some methods, like PCN and 3D-Capsule, and is slightly smaller than the recent approaches, like PF-Net and VRC-Net. 
In our experiment, we train all methods without label information, and calculate the CD loss between the completion result and the ground truth. 

We test all methods in 13 categories, and the results in Table \ref{t1} show that our method has a higher average completion precision than its competitors. Our method achieves the best result in 8 of 13 categories, and ranks the second in 5 of 13 categories (Airplane, Lamp, Laptop, Mug and Skateboard).  Compared with the result of the third-ranked ECG-Net, our method reduces the average CD by 0.37, which is 15.97\% lower than the ECG-Net's result (2.316 in terms of average). 
ECG-Net also uses graph convolution (ours adopts adaptive graph convolution), and it takes a similar strategy to deal with graph convolution's input like ours. But our SPCNet achieves better results in all categories in terms of CD, which proves the generalization ability of adaptive graph convolution for completing shapes across different categories. 

In Table \ref{t1}, both MSN and PF-Net are typical point cloud completion methods, which utilize a coarse-to-fine strategy and achieve the shape generation with a hierarchical structure. Compared with them, our
SPCNet adopts a similar strategy but receives much better results. The improvements should be attributed to the proposed ACM in SPCNet, which substantially helps to generate points in local regions in a locally structured pattern. 

To visually demonstrate that our SPCNet has better performance than other methods, we choose two compared point cloud completion methods from Table \ref{t1}, and show the results in Fig. \ref{compare}. 
We can observe that SPCNet generates the completion results with much better shape structures. In the table category, the visual result shows that the point distribution on the table's boundary predicted by SPCNet is more uniform and smoother than the other methods. When it comes to the chair, SPCNet generates clearer and more detailed structures of the chair legs. Meanwhile, its competitors may wrongly preserve the spatial structures of the whole model or involve noise at the edges, such as the Airplane and Mug.

\subsection{Ablation Experiments}
To demonstrate the usefulness of each module, we change some key modules and retain the other modules. 
\textit{For simplicity, we do not apply the proposed cycle loss in certain ablation experiments. }

\textbf{Effect of SCM.} 
We respectively take ${P_{M/K^{2}}^{\hat{}}}$, ${P_{M/K}^{\hat{}}}$, ${P_{M}^{\hat{}}}$ and ${P_{M}^{*}}$ to compare with ${P_{M}}$. Table \ref{t2} demonstrates our network's completion result is becoming better by SCM. It should be noted that we only use Loss1 (in Fig. \ref{cycleloss}) to optimize the above networks. 

First, we validate that three SCM modules outperform than 1 or 2 SCM modules. Adding more SCM modules (4 or more) can potentially improve the result, but the improvement may be slight and induce more computation time. For example, the first and second SCMs take low-resolution point clouds as input and have few network parameters. In contrast, the third SCM takes a high-resolution point cloud as input, and thus it contains more network parameters (detailed network structure of all of them is included in the supplementary material). The third SCM is more time-consuming than the previous two SCM modules. If we add the fourth SCM into the network, it should be designed like the third SCM. This may result in almost double time cost of the previous training and testing. Thus, we set the number of SCM to 3.    

\begin{table}[!ht]
\centering
\footnotesize
\begin{tabular}{ccccc}
\toprule[1pt]
Category             & ${P_{M/K^{2}}^{\hat{}}}$  & ${P_{M/K}^{\hat{}}}$   & ${P_{M}^{\hat{}}}$        & ${P_{M}^{*}}$              \\
\hline\hline
Airplane             & 7.062                     & 5.155                  & 1.981                     & \textbf{1.954}                  \\
Bag                  & 23.51                     & 12.43                  & 8.897                     & \textbf{8.827}                  \\
Cap                  & 29.26                     & 18.78                  & 7.465                     & \textbf{7.372}         \\
Car                  & 14.14                     & 8.992                  & 5.099                     & \textbf{4.803}         \\
Chair                & 12.29                     & 7.329                  & 4.108                     & \textbf{3.919}         \\
Guitar               & 3.271                     & 1.817                  & 1.033                     & \textbf{0.781}                  \\
Lamp                 & 20.32                     & 17.23                  & 7.589                     & \textbf{7.038}         \\
Laptop               & 9.175                     & 4.546                  & 2.598                     & \textbf{2.596}         \\
Motorbike            & 10.43                     & 6.473                  & 4.123                     & \textbf{3.991}         \\
Mug                  & 23.16                     & 10.76                  & 6.357                     & \textbf{6.312}         \\
Pistol               & 7.297                     & 4.955                  & 2.328                     & \textbf{2.227}         \\
Skateboard           & 7.823                     & 3.131                  & 2.107                     & \textbf{1.979}         \\
Table                & 13.18                     & 6.680                  & 4.005                     & \textbf{3.835}         \\
\hline
Mean                 & 13.91                     & 8.329                  & 4.438                     & \textbf{4.279}         \\
\bottomrule[1pt]
\end{tabular}
\vspace{2pt}
\caption{Effect of SCM (stepwise completion module). We calculate the CD loss of 13 categories from the test set, and the last row denotes the mean CD loss of these categories. We calculate these results by missing parts between them with $P_M$, scaled by 1000.
}
\label{t2}
\end{table}

\textbf{Effect of VMLP.} 
We name our intact network as Whole-Net, and change the structure of VMLP in our Whole-Net to get some different networks:\\ 
(1) Replacing VMLP with the standard PointNet-MLP that has the same dimension and kernels but without maxpooling and
concatenation operations of the last four layers (PointNet-MLP).\\ (2) Replacing the VMLP's three MLP-Sub-Net to one, which has triple dimensions and kernels than before (One Sub-Net).

The detailed information of PointNet-MLP and One Sub-Net can be found in the supplementary material.
We only use Loss1 (in Fig. \ref{cycleloss}) to optimize the above networks. 
Table \ref{t3} shows the results. 
Obviously we will receive worse results if any module is changed, and therefore we can confirm the necessity of the whole VMLP.
We can demonstrate that our VMLP can extract a better global feature than the PointNet's original encoder. Furthermore, we can see that our three MLP-Sub-Net structure can better extract spatial geometrical information from (x,y,z) coordinates. 

\begin{table}[!ht]
\centering
\begin{tabular}{ccccccc}
\toprule[1pt]
Category        & PointNet-MLP       & One Sub-Net            & Whole-Net(1L)            \\
\hline\hline
Airplane        & 1.125             & 1.092                      & \textbf{1.047}       \\
Chair          & 1.904             & 1.907                      & \textbf{1.898}       \\
Table           & 2.097             & 2.058                      & \textbf{1.904}       \\
\hline
Mean            & 1.708             & 1.685                      & \textbf{1.617}       \\
\bottomrule[1pt]
\end{tabular}
\vspace{2pt}
\caption{Ablation results of VMLP in our SPCNet. We calculate these results with the whole model between ${P_N}\cup{P_{M}^{*}}$ and ${P_N}\cup{P_M}$, scaled by 1000.
}
\label{t3}
\end{table}

\textbf{Effect of feature aggregation and ACM.}
We name our intact network as Whole-Net, and change the structure of ACM in our Whole-Net, and get some different networks:\\ (1) Without the feature aggregation operation before feeding the feature into ACM (without-agg).\\ 
(2) Replacing adaptive graph convolution with graph convolution in \cite{9093117} that has same dimension and kernels (without-adapt). 

We only use Loss1 (in Fig. \ref{cycleloss}) to optimize the above networks. 
Table \ref{t4} shows the results. 
Obviously we will receive worse results if any module is changed, and therefore we can confirm the necessity of feature aggregation and the whole ACM.
We can demonstrate that our ACM can extract a better local feature than the original graph convolution. Furthermore, we can see that our feature aggregation can improve the quality of current global features with the last sparse global feature. 

\begin{table}[!ht]
\centering
\begin{tabular}{ccccccc}
\toprule[1pt]
Category        & without-agg       & without-adapt            & Whole-Net(1L)            \\
\hline\hline
Airplane        & 1.132             & 1.195                      & \textbf{1.047}       \\
Chair           & 1.971             & 2.124                      & \textbf{1.898}       \\
Table           & 2.068             & 2.082                      & \textbf{1.904}       \\
\hline
Mean            & 1.723             & 1.800                      & \textbf{1.617}       \\
\bottomrule[1pt]
\end{tabular}
\vspace{2pt}
\caption{Ablation results of feature aggregation and ACM in our SPCNet, calculated from whole shape and scaled by 1000.
}
\label{t4}
\end{table}

\textbf{Effect of FPS.} 
We name our intact network as Whole, and replace the FPS module in our Whole-Net, and get some different networks (FPS, $P_{N/K}$ and $P_{(N/K)/K}$ are shown in Fig. \ref{net}):\\
(1) Replacing FPS with random point sampling (RPS).\\ 
(2) Replacing $P_{N/K}$ with $P_{N}$ after getting the FPS result (PNK-PN).\\ 
(3) Replacing $P_{(N/K)/K}$ with $P_{N}$ (PNKK-PN).

We only use Loss1 (in Fig. \ref{cycleloss}) to optimize the above networks. 
Table \ref{t5} shows the results. 
We receive worse results if any module is replaced, and thus we can observe the necessity of the FPS module. It demonstrates that our FPS result can expose more global features than both the random sampling result and original input. 

The main reason for adopting the multi-resolution strategy is to extract more information with less time cost. Specifically, when setting the sampling rate to be $K=4$, we may develop a network with $4N$ parameters for the original input, and two networks of $2N$ and $N$ parameters for two kinds of sampled inputs. With the smaller model and a low-resolution input, we can greatly reduce the time cost. 
If we replace the sampled inputs with the original input, the smaller network does not work better, leading to a worse result, which has been demonstrated in the ablation experiment ``Effect of FPS''. If we set all parameters to $4N$ and take all input to be original, the result will be better actually, but it will take nearly three times in time cost for training and testing. 
We have included detailed multi-resolution network architectures with their corresponding network parameters in the supplementary material, including all SCM, VMLP and ACM modules for different-resolution input.

\begin{table}[!ht]
\centering
\begin{tabular}{ccccc}
\toprule[1pt]
Category        & RPS       & PNK-PN        & PNKK-PN       & Whole(1L)             \\
\hline\hline
Airplane        & 1.371     & 1.075         & 1.108         & \textbf{1.047}        \\
Chair           & 2.284     & 1.935         & 1.952         & \textbf{1.898}        \\
Table           & 2.347     & 1.929         & 1.967         & \textbf{1.904}        \\
\hline
Mean            & 2.001     & 1.646         & 1.675         & \textbf{1.617}        \\
\bottomrule[1pt]
\end{tabular}
\vspace{2pt}
\caption{Ablation results of replacing FPS in the whole SPCNet, calculated from whole shape and scaled by 1000.}
\label{t5}
\end{table}

\textbf{Effect of cycle loss.} 
We name our intact network as Whole-Net, and compare the Whole-Net (4L) with two different baselines: one loss network Whole-Net (1L) and two loss network Whole-Net (2L). Recalling Fig. \ref{cycleloss}, 1L only uses Loss1, and 2L uses Loss1 and Loss2.
Actually, for the same number of training epochs, the network with the cycle loss will do more computation than the network without the cycle loss. The cycle loss enhances the results by slowing down the speed of convergence. Here, we take the same but large enough epochs to ensure that all methods are converged finally. 
Table \ref{t6} shows the results of removing the cycle loss, and we can observe the effectiveness of all cycle losses. 

\begin{table}[!ht]
\centering
\begin{tabular}{cccc}
\toprule[1pt]
Category            & 1L                & 2L                    & Whole-Net(4L)           \\
\hline\hline
Airplane            & 1.047             & 1.032                 & \textbf{1.014} \\
Chair               & 1.898             & 1.890                 & \textbf{1.835} \\
Table               & 1.904             & 1.897                 & \textbf{1.867} \\
\hline
Mean                & 1.617             & 1.606                 & \textbf{1.572} \\
\bottomrule[1pt]
\end{tabular}
\vspace{2pt}
\caption{Ablation results of the cycle loss, calculated from whole shape and scaled by 1000. }
\label{t6}
\end{table}

\begin{table*}[!ht]
\centering
\footnotesize
\begin{tabular}{ccccccc}
\toprule[1pt]
Category             & 75\%(1L)        & 50\%(1L)          & 25\%(1L)   & 75\%(whole)       & 50\%(whole)            & 25\%(whole)    \\
\hline\hline
Airplane             & 1.541           & 1.047             & 0.641      & 1.439             & 1.014                  & 0.629          \\
Bag                  & 6.558           & 3.491             & 1.949      & 6.414             & 3.428                  & 1.922          \\
Cap                  & 7.409           & 3.276             & 1.992      & 7.255             & 3.267                  & 1.971          \\
Car                  & 3.545           & 2.312             & 1.244      & 3.432             & 2.281                  & 1.226          \\
Chair                & 3.363           & 1.898             & 1.077      & 3.237             & 1.835                  & 1.055          \\
Guitar               & 0.692           & 0.447             & 0.285      & 0.616             & 0.392                  & 0.269          \\
Lamp                 & 5.687           & 2.723             & 1.693      & 5.586             & 2.714                  & 1.679          \\
Laptop               & 2.250           & 1.358             & 0.748      & 2.113             & 1.306                  & 0.736          \\
Motorbike            & 3.159           & 1.995             & 1.207      & 3.050             & 1.893                  & 1.181          \\
Mug                  & 5.194           & 2.983             & 1.981      & 5.092             & 2.976                  & 1.960          \\
Pistol               & 1.972           & 1.146             & 0.730      & 1.838             & 1.124                  & 0.723          \\
Skateboard           & 1.556           & 1.247             & 0.789      & 1.419             & 1.206                  & 0.778          \\
Table                & 3.413           & 1.904             & 1.216      & 3.294             & 1.867                  & 1.198          \\
\hline
Mean                 & 3.564           & 1.987             & 1.196      & 3.455             & 1.946                  & 1.179          \\
\bottomrule[1pt]
\end{tabular}
\vspace{2pt}
\caption{Robustness test. We calculate the CD loss of 13 categories from the test set, and the last row denotes the mean CD loss of these categories. 
The left three colomuns are trained without the cycle losses, and the right three columns are trained with the cycle losses. }
\label{t7}
\end{table*}

\subsection{Robustness}
To demonstrate the robustness of our SPCNet, we test our network and training strategy in varying incomplete inputs. 
Here, we train and test our SPCNet again with missing rates of 25\% and 75\%, respectively.  
First, we train and test the network without the cycle losses. Table \ref{t7} shows the results for the missing rates of 25\%, 50\% and 75\%, respectively. We can obviously observe that the completion effect becomes worse with the increase of missing degrees.

\begin{figure}[!ht]
\centering
\includegraphics[width=1\columnwidth]{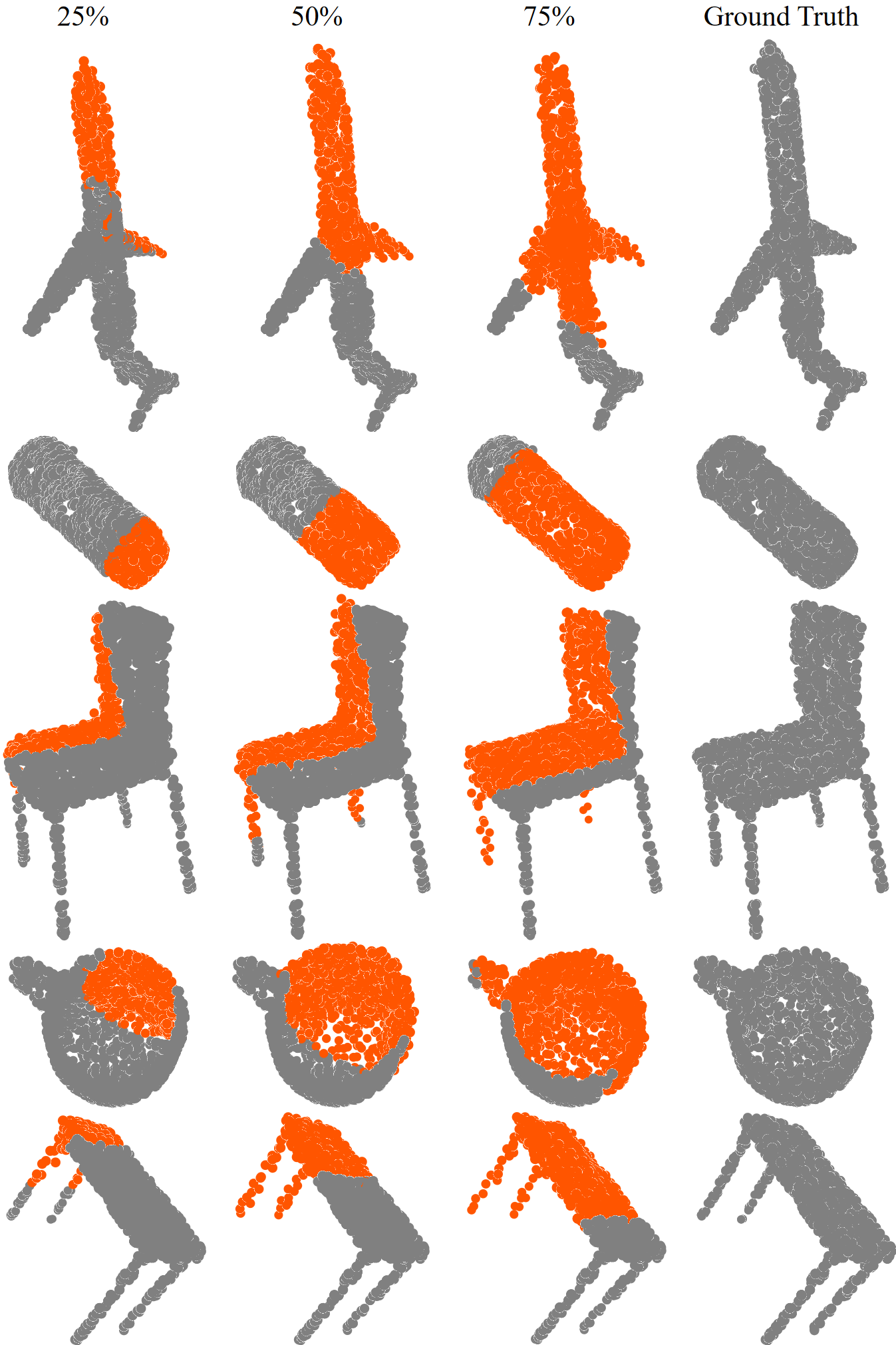} 
\caption{Completion results under different missing degrees (25\%, 50\%, 75\% missing degrees, respectively). }
\label{robust}
\end{figure}

Then, we jointly train the networks in 25\% and 75\% missing degrees with the cycle losses. In Fig. \ref{cycleloss}, the orange arrow denotes the network in a 25\% missing degree, and the red arrow denotes the network in a 75\% missing degree, but they are all the same when we train the network in a 50\% missing degree which can reduce the number of model parameters. Table \ref{t7} also shows the results of our training strategy, which verifies the robustness of our method. Fig. \ref{robust} includes the visual results that show our SPCNet can generate outstanding completion results even in a 75\% missing degree. This further demonstrates the robustness of our SPCNet. 

\section{Conclusion}
We proposed an effective stepwise point cloud completion network (SPCNet) for various 3D models with large missings. SPCNet has a hierarchical bottom-to-up network architecture. 
We firstly receive different scales of representations from down-sampling, and each scale of representation involves its global and local geometric information at that scale. Then, we utilize these representations to generate progressive results by SCM. We finally formulate a new cycle loss to enhance the generalization of our network. 
We demonstrate the effectiveness, generalization and robustness of our method with extensive experiments. 
Our method can reap a complete, uniform and dense result from an incomplete shape. This will facilitate downstream tasks such as object detection and segmentation. 

\section*{Acknowledgments}
This work was supported by the National Natural Science Foundation of China (No. 62172218, No. 62032011), the Free Exploration of Basic Research Project, Local Science and Technology Development Fund Guided by the Central Government of China (No. 2021Szvup060), and the Natural Science Foundation of Guangdong Province (No. 2022A1515010170), and a grant from the Research Grants Council of the Hong Kong Special Administrative Region, China (No. UGC/FDS16/E14/21).

\bibliographystyle{eg-alpha-doi} 
\bibliography{egbibsample}       

\end{document}